\documentclass{article}

\usepackage{microtype}
\usepackage{graphicx}
\usepackage{subfigure}
\usepackage{booktabs} 
\usepackage{svg}
\usepackage{multirow}
\usepackage{makecell}

\PassOptionsToPackage{hyphens}{url}\usepackage{hyperref}

\usepackage[accepted]{icml2024}

\usepackage{amsmath}
\usepackage{amssymb}
\usepackage{mathtools}
\usepackage{amsthm}

\usepackage[capitalize,noabbrev]{cleveref}

\theoremstyle{plain}

\theoremstyle{definition}

\theoremstyle{remark}

\usepackage[textsize=tiny]{todonotes}

\begin{document}

\twocolumn[
\icmltitle{Conditional Language Learning with Context}



\icmlsetsymbol{equal}{*}

\begin{icmlauthorlist}
\icmlauthor{Xiao Zhang}{yyy}
\icmlauthor{Miao Li}{yyy}
\icmlauthor{Ji Wu}{yyy,xxx}
\end{icmlauthorlist}

\icmlaffiliation{yyy}{Department of Electronics Engineering, Tsinghua University}
\icmlaffiliation{xxx}{College of AI, Tsinghua University}

\icmlcorrespondingauthor{Ji Wu}{wuji\_ee@mail.tsinghua.edu.cn}

\icmlkeywords{Machine Learning, ICML}

\vskip 0.3in
]



\printAffiliationsAndNotice{}  

\begin{abstract}
    Language models can learn sophisticated language understanding skills from fitting raw text. They also unselectively learn useless corpus statistics and biases, especially during finetuning on domain-specific corpora. In this paper, we propose a simple modification to causal language modeling called conditional finetuning, which performs language modeling conditioned on a context. We show that a context can ``explain away" certain corpus statistics and make the model avoid learning them. In this fashion, conditional finetuning achieves selective learning from a corpus, learning knowledge useful for downstream tasks while avoiding learning useless  corpus statistics like topic biases. This selective learning effect leads to less forgetting and better stability-plasticity tradeoff in domain finetuning, potentially benefitting lifelong learning with language models.
\end{abstract}

\section{Introduction}
\label{introduction}

Language models pretrained on large-scale corpus have shown impressive performance on a wide variety of downstream tasks \citep{flan, llama, gpt4}. 
It is impressive that these language models learn sophisticated knowledge and reasoning abilities solely from training on raw text with a causal language modeling objective. The objective is also used when adapting the pretrained general-purpose language models to specific domains, via finetuning on a domain corpus (also called ``continual pretraining'') \citep{codex, minerva, medpalm2}. 

Although finetuning effectively improves the model's domain knowledge and performance on domain tasks, it can also lead to forgetting of existing knowledge \citep{continuallmtask3, continuallmknowledge} due to modifying the pretrained model. 
It is also observed that finetuning can lead to over-adaptation to the statistical properties of the domain corpus, causing the model to be biased heavily towards certain topics and styles \citep{dissectlm}.

In domain finetuning, it would be desirable to improve the model's domain knowledge without learning useless statistics and biases from the corpus. The causal language modeling objective maximizes likelihood of all the tokens in the corpus, and is therefore unselective in what kind of information to learn from the corpus. In this paper, we propose a simple enhancement to causal language modeling called conditional finetuning, that uses contexts to achieve selective learning of useful information from the corpus.

It is well-known that the behavior of pretrained language models is sensitive to contextual information in the input during inference. For example, few-shot prompting can let models learn to perform tasks based on examples in the context \cite{iclsurvey}. Specific instructions like chain-of-thought \cite{cot} and self-verification \cite{selfverification} could guide the model towards certain behaviors like multi-step reasoning. For dialog and assistant use cases, language models can be further finetuned on instruction-following data to make them more sensitive to instructions in the context \cite{instructgpt, flan, t0}. While the effect of context during inference has been extensively studied, its role during the pretraining phase is less explored. In this paper, we investigate how adding a context to language modeling could affect the model's learning behavior during pretraining and domain finetuning.

The two main contributions of the paper are:

\begin{figure*}[t]
    \centering
    \includegraphics[width=0.75\linewidth]{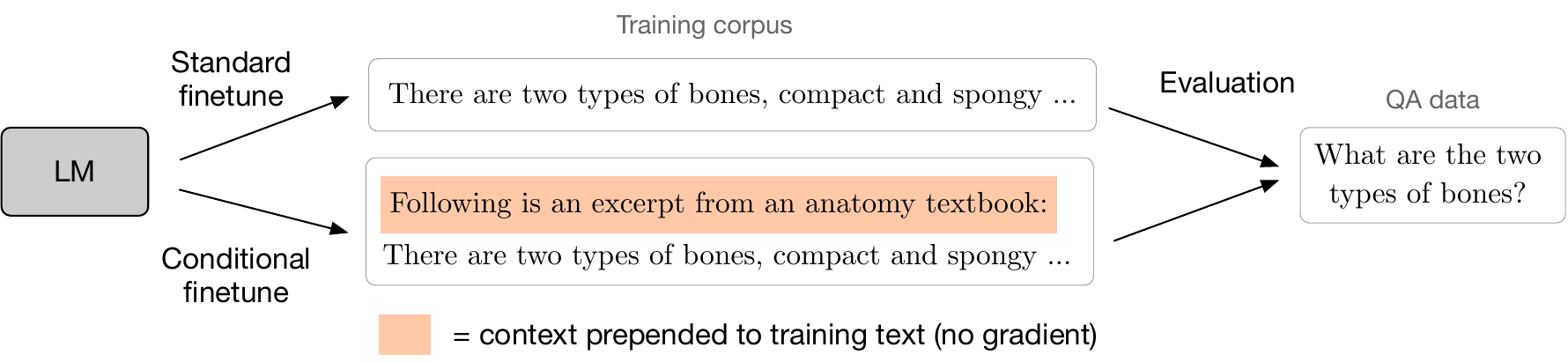}
    \caption{Illustration of conditional finetuning a language model on domain corpus. Compared to standard finetuning, conditional finetuning prepends a context to each document and only learns information conditioned on the context.}
    \label{fig:overall}
\end{figure*}

We propose conditional finetuning, a domain finetuning method for language models that adds a context to causal language modeling. We reveal how adding a context affects the language modeling objective.
In conditional finetuning, we use a piece of text as context and prepend it to corpus text during finetuning. The idea of the method is illustrated in \cref{fig:overall}. We show that the context can ``explain away'' statistical properties of the corpus so that the model would ignore them and avoid learning them in finetuning. For example, when finetuning on a domain corpus with a domain hint as context, the model can keep its topic prior almost unchanged, without adapting to the topic distribution of the domain corpus, as typically observed in conventional finetuning.

We also show that conditional finetuning achieves selective learning of useful information. Knowledge useful for downstream tasks is learned without compromise, while learning of corpus statistics is reduced, which leads to significantly less modification to the pretrained model during finetuning. This selective learning leads to better stability-plasticity tradeoff and less forgetting in one-time finetuning (transfer learning) and multiple-time finetuning (continual learning) scenarios, making it a better alternative to conventional finetuning for lifelong learning with language models.

We release our code implementation along with the original part of the data used in the paper\footnote{\url{https://github.com/xiaozeroone/conditional_finetune}}.

\section{Related Work}

\paragraph*{Domain finetuning of language models.} Finetuning pretrained language models on domain corpus is a common approach to enhances domain knowledge and help language models perform better on domain tasks, for example, in mathematics \citep{minerva}, coding \cite{codex} and medicine \cite{medpalm2}. Finetuning on multiple corpora can help language models continually learn knowledge from multiple domains \citep{rewarmup, continualpretraining, continualpretraining2}, as a form of continual or lifelong learning \citep{lifelonglearning}. Besides learning knowledge, finetuning can also lead to over-adaptation to the statistics of domain corpus like topic and style, as an unwanted side effect \cite{shortcutlm, dissectlm}.


\paragraph*{Inferencing with context: in-context learning.} With the extensive use of autoregressive language models for question answering and problem-solving, contexts (prompts) are often used to provide extra information or guidance to the model. For example, In-context learning \cite{iclsurvey} provide a few examples as context for the model to learn from during inference. Chain-of-thought \cite{cot} and tree-of-thought \cite{tot} use special instructions to guide the model to perform step-wise reasoning. Self-verification \cite{selfverification} also uses the model's own prediction as context to perform an extra verification step to improve the accuracy of reasoning.

\paragraph*{Learning context for inference: prompt learning.} Because context can significantly affect language model's performance in inference, prompt learning methods are used to learn optimal contexts for a target task. 
Prompt tuning \cite{prompttuning} and prefix tuning \cite{prefixtuning} optimize soft prompts for target tasks as a parameter-efficient tuning alternative to full finetuning. For better expressivity, soft prompts can be individually given to each layer of the model \cite{multilayerprefixtuning}. Gradient-free black-box optimization \cite{blackboxtuning} can also be used to learn prompts where models weights are inaccessible, such as when using commercial models.

\paragraph*{Training on context: instruction finetuning.} To enhance language model's ability to use context, instruction finetuning \cite{instructgpt} finetunes model on instruction-response pairs to make model better at following instructions. Training with diverse instructions, e.g., FLAN \cite{flan} and T0 \cite{t0}, helps model generalize to new instructions and tasks. Training with few-shot and chain-of-thought examples also helps model better utilize those kinds of context in inference \cite{flanv2}. 

Such methods usually train with loss on the ``response'' part of the instruction-response pairs, using the instruction as context. The goal is to learn the relationship between various instructions and their corresponding responses for better instruction following. By contrast, conditional finetuning is used in continual pretraining where the goal is to learn knowledge from corpora. Conditional finetuning does not learn information from the context, but instead uses the context to ``explain away" corpus statistics thus reduce learning of those useless statistics.

\paragraph*{Selective learning.} Generative models are trained to directly maximize the likelihood of data, so they tend to indiscriminately learn all patterns within the data. To make models selectively learn certain patterns from data, one can perform data selection \cite{dataselectiontransfer,dataselectlm}, choosing subsets of training data that contains the desired patterns, or perform soft example selection by importance sampling \cite{importancesamplingdl}. For language models, one can also use loss re-weighting at token level to selectively learn from informative tokens \cite{tokenmask}. Attention guidance is another approach that leverages the mechanistic interpretability of attention to make model ``focus'' more on certain features in the input, thus selectively learn certain features more \cite{salientattention, attentionmod, topdownattention}. More related to our approach, it is possible to learn an ensemble of models in order to factor learned patterns into different models. This is successfully used to de-bias models in natural language tasks \cite{productofexperts, productofexperts3, productofexperts2}.

\section{Conditional Learning}
\label{sec:conditional_learning}
We first use the language of probabilistic modeling to illustrate the idea of conditional learning.

Consider a probabilistic model $p$ of some data.
Suppose an example $x$ has a property $c$ that can be inferred from $x$, i.e., $p(c|x) = 1$. Then the probability of $x$ can be decomposed as 
\begin{align}
    p(x) = p(x,c) = p(x|c)p(c).
    \label{eq:decompose}
\end{align}
If we want to fit the model on data, increasing the likelihood of $x$ under $p$, we can either increase $p(c)$ or $p(x|c)$. The former is fitting to the property $c$. The latter is leaning the regularities in $x$ \textit{besides} the property $c$, which we refer to as \textit{conditional learning} in this paper.


If a set of examples $\{x_i\}_{i=1}^N$ all have the same property $c$, then the average log-likelihood of the dataset is
\begin{align}
    \frac{1}{N}\sum_i \log p(x_i) = \frac{1}{N}\sum_i \log p(x_i|c) + \log p(c).
    \label{eq:log_likelihood}
\end{align}
When fitting model $p$ to the dataset by maximizing data likelihood,  according to \cref{eq:log_likelihood}, increasing $\log p(c)$ will likely increase data likelihood faster than increasing $\log p(x_i|c)$ for certain individual examples. This implies that the model could be biased towards adapting to the common property $c$ of the data if such property is present. Moreover, if the property $c$ is simple, the model may also adapt $p(c)$ faster and earlier than learning $p(x|c)$ \cite{shortcut, shortcutlm}.


In this paper, we specifically explore this situation in language modeling. 
When finetuning a general-purpose language model on a specialized domain corpus, the model can exhibit a noticeable bias towards the domain. This bias arises because the domain acts as a common property among the corpus documents, leading the model to significantly adapt its topic prior in favor of the domain topic \cite{dissectlm}.
Nonetheless, the ultimate goal of finetuning language models is to enhance their domain-specific knowledge without compromising their general knowledge and ability \cite{continuallmtask3,continuallmknowledge}. 

Knowledge is often embedded in text in the form of conditional token probabilities like 
$p(x_{[k...n]}|x_{[1...k]})$. For instance, $p(\text{``\textit{London}"}|\text{``\textit{The capital city of England is}"})$ can represent the factual knowledge within the sentence ``The capital city of England is London".


In this case, learning the conditional probability $p(x_{[k...n]}|x_{[1...k]},c)$ conditioned on corpus-level properties $c$ (such as the topic of the corpus) is enough for the purpose of knowledge learning. i.e., learning $p(x|c)$ in \cref{eq:decompose} is sufficient in domain finetuning of language models.
Learning $p(\text{topic})$ and other corpus properties are not necessary and may be harmful in case of lifelong learning \cite{lifelonglearning} because they introduce unnecessary bias.

Luckily, it is straightforward to perform conditional learning in causal language modeling, where $p(x)$ is decomposed as probabilities of each token given the previous tokens:
\begin{align}
    p(x) = \prod_{i=1}^n p(x_i|x_{<i}).
    \label{eq:causal_lm}
\end{align}
Now we can explicitly prepend the property $c$ (in text form) to the original text $x$. Here, we use notation $\langle,\rangle$ to indicate concatenation of text. Under causal language modeling, applying \cref{eq:causal_lm} gives the decomposition
\begin{align}
    p(\langle c, x\rangle) = p(x|c)p(c).
\end{align}
To only learn the conditional probability $p(x|c)$, we feed the concatenated sequence $\langle c, x\rangle$ into the model and use the negative log-likelihood on tokens in $x$ as loss, which directly corresponds to $p(x|c)$.
It can be implemented on standard causal language modeling by simply masking out the loss on the first few tokens corresponding to context $c$.

In theory, we could prepend any context $a$ to the text, not necessarily those representing properties of the corpus. In this case, we will have
\begin{align}
    p(\langle a, x\rangle) = p(x|a)p(a) = p(x|c,a)p(c|a)p(a)
\end{align}
we will see later from experiments that supervising $p(x|a)$ will lead to learning both the conditional probability $p(x|c,a)$ and a ``conditional prior" $p(c|a)$, such as a conditional topic prior $p(\text{topic}|a)$ activating domain topics when the context $a$ is given.

\paragraph*{Selective learning.} Conditional learning is a form of selective learning: it factorizes the objective in \cref{eq:log_likelihood} into two parts and learns one part of it. It turns the unselective learning objective of language modeling into a selective learning objective that only learns knowledge useful for downstream tasks and avoids learning corpus statistics, reducing side effects such as over-adaptation and forgetting.

\section{Conditional Language Modeling with Context}
\label{sec:conditional_lm}

In this section, we apply conditional learning to language modeling and finetune a pretrained language model on a domain corpus. We analyze the effect of conditional learning and show that it reduces the learning of the topic prior $p(\text{topic})$. Instead, the model uses a conditional prior $p(\text{topic}|c)$ to fit the topic distribution of the domain corpus.

\paragraph*{Data and model.} We use the medical textbooks provided with the MedQA dataset \cite{medqa} as a domain corpus to finetune LLaMA-2 \cite{llama2}, a series of state-of-the-art language models pretrained on large-scale general text. The medical textbooks collection contains 18 textbooks on various medical subjects and has a total of 25.7M tokens. The textbooks are dense in medical knowledge and are a good candidate for studying domain-specific knowledge learning.

We finetune the model with the AdamW optimizer \cite{adamw}, a learning rate of 3e-5, and a batch size of 16. The maximum sequence length is set to 2048.
We use the Transformers library \cite{hf} and an NVIDIA A100 GPU for the experiments.

\paragraph*{Conditional finetuning.} In conditional finetuning, we prepend a context $c$ to each document $x$ and use $\langle c, x\rangle$ instead of $x$ as training examples. The training objective is the same as conventional causal language modeling,  except that the loss from the first $|c|$ tokens are ignored ($|c|$ = the number of tokens in $c$). Conditional finetuning uses the same hyperparameters as standard finetuning.

The context $c$ in only used in training. During inference, conditionally finetuned models are used \textit{without} the context just as normal language models.

\subsection{Conditional Finetuning Reduces Learning of the Topic Prior $p(\text{topic})$} 

To examine the effect of conditional finetuning, we first use a short sentence hinting the topic as the context, i.e.,  $c=$``\textit{Following is an excerpt from a medical textbook.}"


To measure the topic prior of a language model, we use a simple topic likelihood probe. The probe is a sentence ``\textit{The text is about [topic]}." where [topic] is replaced by a topic word. The likelihood of the topic word $p(\textit{[topic]}|\textit{``The text is about"})$ given by the model indicates the model's topic prior.

\begin{figure}[h]
    \centering
    \includesvg[width=\linewidth]{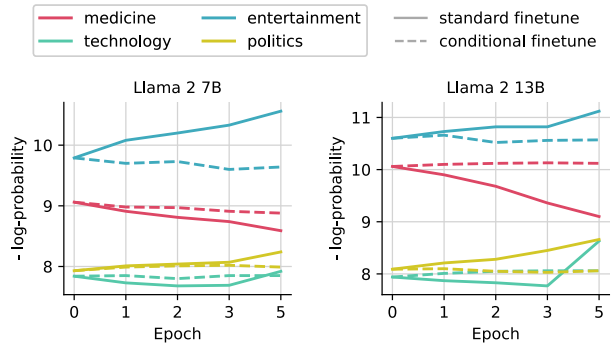}
    \vskip -0.1in
    \caption{Topic likelihood changes during finetuning. Unlike standard finetuning, conditional finetuning does not significantly change topic likelihoods.}
    \label{fig:topic_prior}
\end{figure}

\begin{figure*}[ht]
    \centering
    \includegraphics[width=\linewidth]{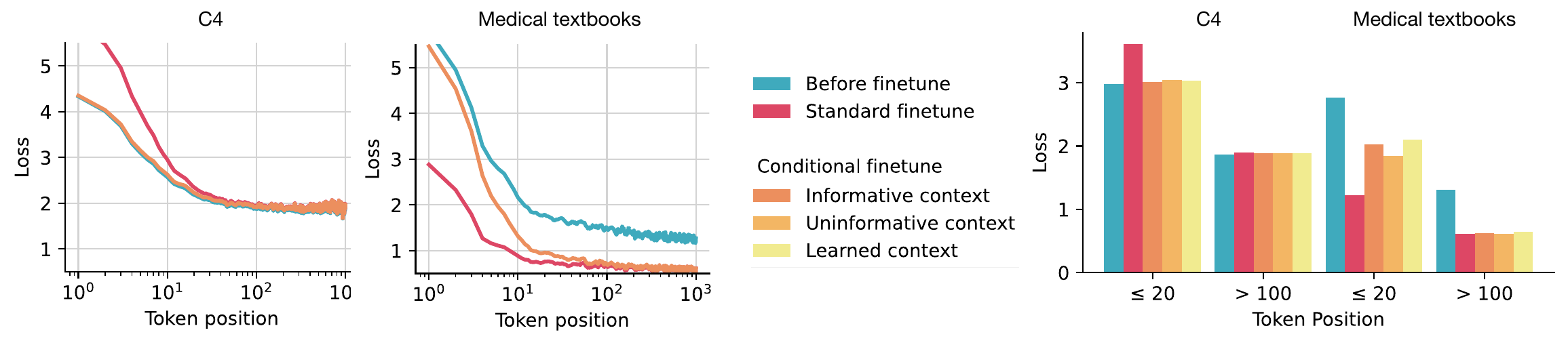}
    \vskip -0.1in
    \caption{Comparing language modeling loss at different token positions, for models finetuned with standard finetuning and conditional finetuning with three types of context. 
    Unlike standard finetuning, conditional finetuning barely increases loss on general text (C4), regardless of the type of context used in training. (model: LLaMA-2 7B)}
    \label{fig:loss_per_position}
\end{figure*}

\cref{fig:topic_prior} shows that topic likelihoods changes significantly during standard finetuning. Finetuning typically increases the likelihood of the domain topic and decreases the likelihood of other topics. Conditional finetuning, on the other hand, keeps the topic likelihoods stable even after several epochs of training. 
This suggests that when topic information is already given as context, the model no longer need to fit its topic prior to the training text, confirming the conditional learning hypothesis in Section \ref{sec:conditional_learning}.

Next, we show how conditional finetuning affects the modeling of a longer piece of text. We calculate the finetuned models' average language modeling loss on medical textbooks and C4 \cite{t5}, a corpus of general web text. We calculated average loss separately at each token position, in a similar fashion as in \citet{dissectlm}.

\cref{fig:loss_per_position} (left) shows that standard finetuning significantly changes loss on the first few tokens of text. Loss decreases on the training corpus and significantly increases on C4. This suggests that the finetuned model becomes highly predisposed to medical texts from the beginning, likely due to an over-adaptation of the topic prior.
On the other hand, conditional finetuning results in a negligible change in loss across all positions on the C4 corpus. This ability to keep loss stable on general corpus suggests a potential to mitigate issues related to over-adaptation and forgetting in finetuning.


\subsection{Conditional Finetuning Learns a Conditional Topic Prior $p(\text{topic}|a)$ regardless of Context $a$} 
Interestingly, we found that conditional finetuning always learns a conditional topic prior $p(\text{topic}|a)$ instead of an unconditional topic prior $p(\text{topic})$, regardless of the context $a$ given in training. To see this, we experiment with three types of context:

\textbf{\textit{Domain hint}} (informative context): A short sentence hinting the topic: ``\textit{Following is an excerpt from a medical textbook.}"

\textbf{\textit{Random}} (uninformative context): A randomly generated Universally Unique Identifier (UUID) string: ``\textit{7a1d64b1-fa43-47a8-9389-60406eb96778.}"

\textbf{\textit{Learned}} (learned context): A soft prompt of 10 vectors learned using the prompt tuning method \cite{prompttuning}. The soft prompt is learned by fitting the training corpus while keeping the language model fixed. The soft prompt is then used as a fixed context during finetuning the whole model. Detailed procedure is described in \cref{app:prompt_tuning}.

Specifically, the learned context is a context that maximizes the likelihood of the training corpus. Therefore, it likely encodes some overall statistics of the corpus, such as the main topic. The context is unlikely to encode detailed knowledge due to the limited capacity of the soft prompt.



\cref{fig:loss_per_position} (right) compares the loss at different token positions of models finetuned with different context types. All types of contexts are similarly effective at keeping loss unchanged on general corpus (C4). This shows that the presence of any context would reduce learning of the topic prior $p(\text{topic})$ regardless of the content of the context.

The reason is that even when provided with a non-informative context $a$, the model learns a conditional topic prior $p(\text{topic}|a)$ that makes $a$ function similarly as an informative context. We can clearly see this from some examples.


\begin{figure}[h]
    \centering
    \includegraphics[width=\linewidth]{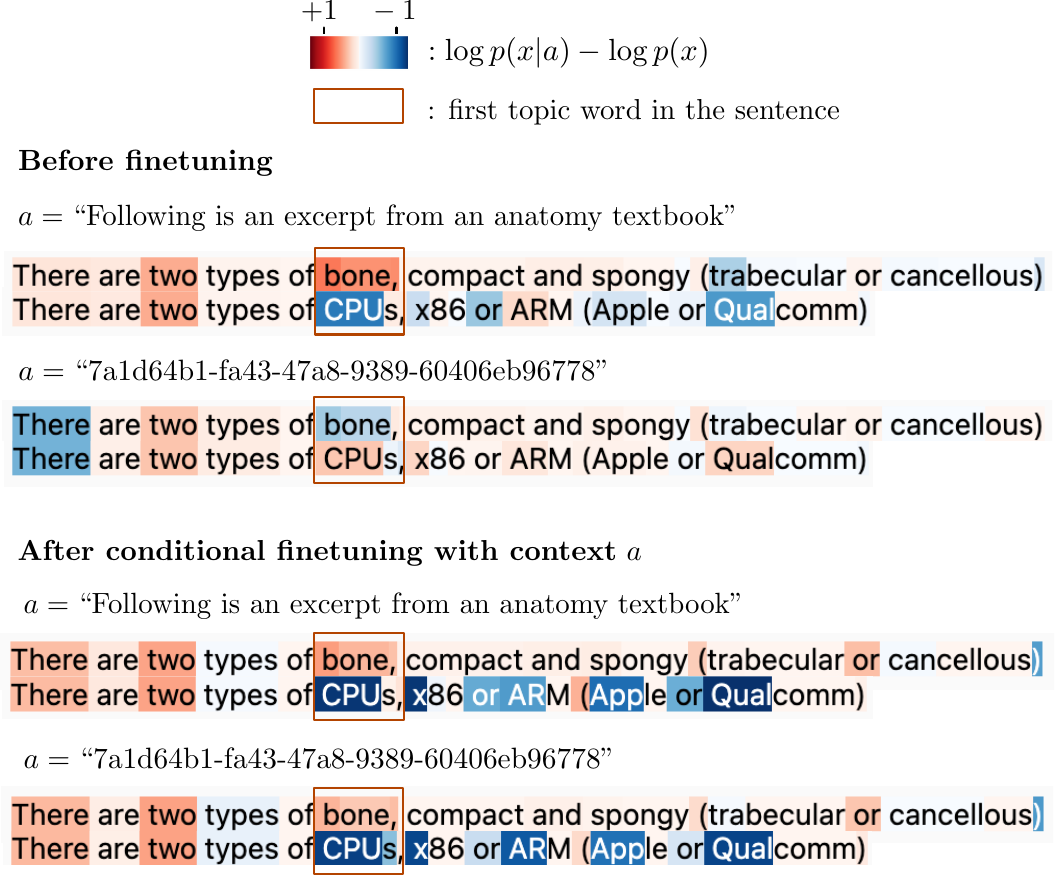}
    \caption{Examples showing the loss change $\log p(x) - \log p(x|a)$ caused by the context. Before finetuning, domain hint makes the model favor the medical term ``bone" while the random UUID string favors technological term ``CPU". After finetuning, both contexts have similar effect of favoring medical terms.}
    \label{fig:context_examples}
\end{figure}

\cref{fig:context_examples} shows an example of the context's effect on the model's prediction. A medical text and a technological text are used as examples. On the original pretrained model, domain hint increases the likelihood of medical terms and decreases the likelihood of technological terms. The random UUID string has the opposite effect, likely because UUID strings are more commonly associated with technological contexts. Despite behaving differently on the pretrained model, all kinds of contexts have a remarkably similar effect on their own conditionally finetuned models. They all make the model strongly favor medical over technological words. This indicates that the model learns a conditional topic prior $p(\text{topic}|a)$ that makes $a$ function similarly as a domain hint. If $a$ is already informative about the topic, $p(\text{topic}|a)$ may simply be reinforced during conditional finetuning.

\subsection{Conditional Finetuning does not Affect Knowledge Learning} 
\cref{fig:loss_per_position} shows that, for later token positions of the training corpus (e.g., $>$100), conditional finetuning achieves the same loss reduction as standard finetuning. This suggests that conditional finetuning likely will not affect the learning of factual knowledge, which is mostly in the form of $p(x_{[k...n]}|x_{[1...k]})$ (discussed in \cref{sec:conditional_learning}). The hypothesis is verified in the next section where we evaluate performance on downstream tasks. 
These findings indicate that conditional training is effective at selective learning, learning knowledge useful for future tasks while avoiding learning corpus statistics that are not useful.


\section{Less Forgetting through Selective Learning}
\label{sec:less_forgetting}


Evidence from language modeling suggests that conditional finetuning performs selective learning. In this section, we further elucidate this selective learning effect. Conditional finetuning modifies the pretrained model less than standard finetuning, and therefore helps achieve less forgetting in transfer learning and continual learning scenarios. At the same time, knowledge learning is not affected.


\subsection{Conditional Finetuning Modifies Model Less}

We use two metrics to measure how much the pretrained language model is modified during finetuning. To estimate the influence of the training objectives on the model, we calculate the gradient norm of standard and conditional finetuning objectives on the pretrained model. 
L2-norm of the gradient is calculated over the training corpus by flattening all parameters of the model into a single vector\footnote{Scaling parameters in layer normalization \cite{layernorm} are excluded as they can have large gradients and dominate the L2 norm when included.}. As shown in \cref{tab:gradient_norm}, conditional training objective has a significantly smaller gradient norm than the standard finetuning objective, even though they both use the same cross-entropy loss on the same training tokens.
This indicates that the conditional finetuning objective requires less modification to the model parameters, likely by removing the gradient for fitting corpus statistics.

\begin{table}[h]
    \centering
    \small
    \setlength\tabcolsep{3.5pt}
    \begin{tabular}{lcc}
    \toprule
    \makecell[l]{\textbf{Gradient norm}\\\textbf{of objective}}  & Standard finetune & \makecell{Conditional finetune\\w/ domain hint}  \\ \midrule
    \textit{On medical textbooks} \\
    LLaMA-2 7B         & 1.93 & 1.13 \\
    LLaMA-2 13B         & 1.56 & 1.06 \\
    \bottomrule
    \end{tabular}
    \caption{Gradient norm of standard finetune and conditional finetune objectives. The conditional finetune objective has a significantly smaller gradient norm.}
    \label{tab:gradient_norm}
\end{table}

To see how much the model changes during finetuning, we can measure the similarity of the model before and after finetuning. We calculate the KL-divergence between the output probability distribution of the pretrained model and the finetuned model, averaged over all tokens. The models are all finetuned for 5 epochs. As shown in \cref{tab:kl_divergence}, on C4, models finetuned with conditional finetuning have significantly smaller KL-divergence to the pretrained model than models finetuned with standard finetuning. This confirms that conditional training modifies the model less than standard finetuning.

\begin{table}[h]
    \centering
    \small
    \setlength\tabcolsep{3.5pt}
    \begin{tabular}{lcc}
    \toprule
    \makecell[l]{\textbf{KL-divergence}\\\textbf{to pretrained model}}  & Standard finetune & \makecell{Conditional finetune\\w/ domain hint}  \\ \midrule
    \textit{On C4} \\
    LLaMA-2 7B         & 0.082 & 0.036 \\
    LLaMA-2 13B         & 0.116 & 0.071 \\
    \bottomrule
    \end{tabular}
    \caption{KL-divergence from the finetuned model to the pretrained model. Conditional finetuning results in a significantly smaller KL-divergence than standard finetuning.}
    \label{tab:kl_divergence}
\end{table}

\subsection{Conditional Finetuning Reduces Forgetting and Maintains Knowledge Learning in Transfer Learning} 
We next show that because conditional finetuning modifies the model less, it achieves less forgetting in transfer learning (one-time finetuning). Also, knowledge learning is uncompromised in selective learning. As a result, conditional training achieves better stability-plasticity tradeoff over learning new knowledge and retaining existing knowledge, the perennial dilemma in lifelong learning \cite{clreview, clnlpreview}.

We evaluate knowledge learning in finetuned language models with question answering tasks, a common approach in previous work \cite{mmlu, medpalm2}. We finetune language models on two kinds of domain text, one specific domain in medicine (medical textbook) and one general domain (Wikipedia). The finetuned models are then evaluated on the corresponding question answering tasks. The two scenarios are described below:


\begin{itemize}
    \item \textbf{Anatomy.} Training corpus: the anatomy textbook from the MedQA dataset \cite{medqa}. QA data: 500 multiple choice quiz questions on core anatomy concepts in the textbook. Quiz questions are generated automatically with GPT-4 \cite{gpt4}. The procedure to generate quiz questions, including prompts examples are described in \cref{app:qa}. QA performance is evaluated with standard 5-shot prompting.
    \item \textbf{SQuAD (closed-book).} Training corpus: Wikipedia excerpts from the SQuAD dataset \cite{squad}. QA data: questions about facts in the Wikipedia excerpts, also from the SQuAD dataset. We turned the reading comprehension dataset of SQuAD into a closed-book QA, by first finetuning the model on the Wikipedia excerpts and then evaluate it on question answering without giving the excerpts. This closed-book QA setting was previously  used to evaluate knowledge learning in language models \cite{tokenmask}. QA performance is measured using normalized F1 score.
    
    Evaluation details are described in \cref{app:qa_eval}. 
    Results on more datasets are given in \cref{fig:qa_transfer_more} in appendix.
\end{itemize}

\begin{figure}[h]
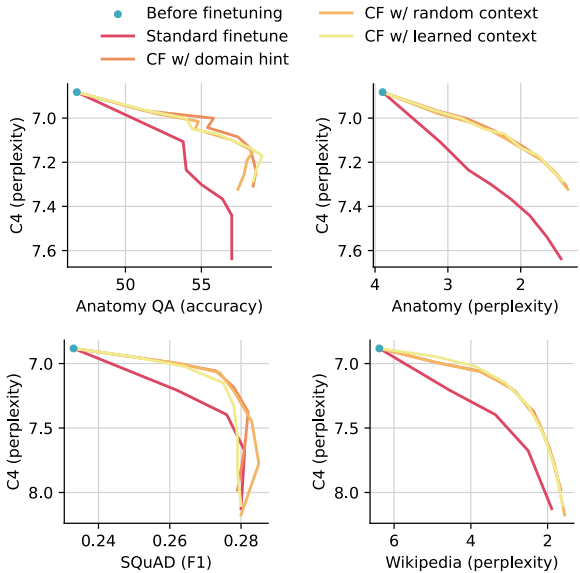

    \centering
    \includesvg[width=0.95\linewidth]{figures/transfer_anatomy.svg}
    \includesvg[width=0.95\linewidth]{figures/transfer_squad.svg}
    \caption{Performance-forgetting tradeoff curve of standard finetuning and conditional finetuning on Anatomy and SQuAD (closed-book).
    Conditional finetuning has less forgetting at similar levels of performance on downstream tasks, achieving significantly better tradeoff than standard finetuning.}
    \label{fig:qa_transfer_learning}
\end{figure}

We plot the stability-plasticity tradeoff curve (or ``performance-forgetting'' curve) of standard finetuning and conditional finetuning in \cref{fig:qa_transfer_learning}. The curves show forgetting as a function of learning. 
Curves on the top-right represent better tradeoff.
To obtain the curve, we finetune the model on domain corpus for different numbers of epochs (from 1 to 8). We evaluate the model on the QA task as a measure of knowledge learning, and use perplexity on C4 as a measure of forgetting of existing information.


\cref{fig:qa_transfer_learning} shows that conditional finetuning achieves significantly better tradeoff than standard finetuning. Furthermore, the tradeoff is relatively insensitive to the type of context.
The maximum achievable performance on QA task is similar for the two training methods, while conditional finetuning has significantly less forgetting at each levels of learning. This shows that by selective learning, conditional finetuning poses less disruption to the information in pretrained model and achieves better stability-plasticity tradeoff.



\subsection{Conditional Finetuning Reduces Forgetting and Improves Knowledge Learning in Continual Learning} 

When continually finetuning on multiple corpora, language models can continually learn new domain knowledge to integrate with existing knowledge. We show that conditional finetuning can also reduce forgetting of previously learned knowledge in a continual learning setting. It results in improved cumulative knowledge learning over the entire course of continual finetuning.

Similar to the transfer learning scenario, we finetune language models on a medical domain and a general domain:
\begin{itemize}
    \item \textbf{Medical textbooks} (13 corpora). Training corpus: 13 medical textbooks from the MedQA dataset \cite{medqa}. Details are described in \cref{app:corpus}. We continually finetune the model on 13 textbooks in the following order: anatomy, biochemistry, cell biology, gynecology, histology, immunology, neurology, obstetrics, pathology, pediatrics, pharmacology, physiology, and psychiatry. QA data: 500 multiple choice quiz questions for each subject, similar to the transfer learning setting (\cref{app:qa}).
    \item \textbf{MRQA (closed-book}, 6 corpora). Training corpus: 6 corpora of Wikipedia pages, web text, and news articles from the 6 reading comprehension datasets in the MRQA benchmark \cite{mrqa}: SQuAD \cite{squad}, NewsQA \cite{newsqa}, TriviaQA \cite{triviaqa}, SearchQA \cite{searchqa}, HotpotQA \cite{hotpotqa}, and Natural Questions \cite{naturalquestions}. 
    We continually finetune the model on corpora provided by each dataset in that order. QA data: questions about facts in the corresponding corpus, also from each of the 6 datasets. Questions are turned into a closed-book format, and performance is evaluated using the evaluation protocol of SQuAD for consistency (\cref{app:qa_eval}).
\end{itemize}


We use the Average Forgetting metric from \citet{agem} to evaluate forgetting in continual learning:
\begin{align}
    F_k = \frac{1}{k-1}\sum_{i=1}^{k-1} \max_{j \in \{1,...,k-1\}} a_{j,i} - a_{k,i}
\end{align}
$F_k$ measures the average forgetting on previous QA tasks after training on the $k$-th corpus.
$a_{k,i}$ is the accuracy on the $i$-th QA task after training on the $k$-th corpus.

We also use Cumulative Accuracy as a measure of the total knowledge learned over the course of continual finetuning:
\begin{align}
    C_k = \frac{1}{n}\sum_{i=1}^n a_{k,i}
\end{align}
$C_k$ measures the average accuracy on all QA tasks after training on the $k$-th corpus\footnote{Compared to Average Accuracy in \citet{agem}, Cumulative Accuracy takes into account the initial performance of pretrained models and is more suited to measure learning on pretrained models.}. $n$ is the total number of tasks.

\begin{table}[h]
    \centering
    \small
    \setlength\tabcolsep{4.8pt}
    \begin{tabular}{lcccc}
    \toprule
    \multirow{2}{*}{\textbf{Performance}}  & \multicolumn{2}{c}{\makecell[c]{Average\qquad \qquad \\Forgetting ($\downarrow$)}} & \multicolumn{2}{c}{\makecell[l]{Cumulative\\Accuracy ($\uparrow$)}} \vspace {0.1cm} \\ 
    & LLaMA-2 7B & 13B & 7B & 13B \\
    \midrule
    \textit{Medical textbooks} \\
    Pretrained          & -    & -  & 53.5 & 59.2 \\
    Standard finetune   & 2.5 & 2.6  & 60.3 & 65.3 \\
    CF (w/ domain hint) & 2.3 & 2.5  & 60.5 & 65.2 \\
    CF (w/ random)      & 2.3 & 2.6  & 60.1 & 65.2 \\
    CF (w/ learned)     & \textbf{2.1} & \textbf{2.3}  & \textbf{60.7} & \textbf{65.6} \\
    \midrule
    \textit{MRQA (closed-book)} \\
    Pretrained       & -    & -  & 0.390 & 0.431 \\
    Standard finetune & 0.026 & 0.014  & 0.390 & 0.449 \\
    CF (w/ random)   & 0.022 & 0.014  & 0.382 & 0.450 \\
    CF (w/ learned)  & \textbf{0.019} & 0.015  & \textbf{0.395} & 0.450 \\
    \bottomrule
    \end{tabular}
    \caption{Continual learning performance of standard finetuning and conditional finetuning (at the last episode, $k$=$n$). CF = Conditional finetune.
    Conditional finetuning has less forgetting and achieves better cumulative accuracy. Learned context has better performance than other types of contexts.
    Note that the metric is F1 instead of accuracy for MRQA.}
    \label{tab:continual_learning}
\end{table}

We adapted the three types of context in \cref{sec:conditional_lm} to use in continual learning: for \textbf{\textit{domain hint}}, we use ``\textit{Following is an excerpt from a [subject] textbook}" as context, where [subject] is replaced by the subject of each textbook. For \textbf{\textit{random}}, we use a different random UUID string for each corpus. For \textbf{\textit{learned}}, we learn soft prompts for each corpus.

\begin{figure}[h]
    \centering
    \includegraphics[width=\linewidth]{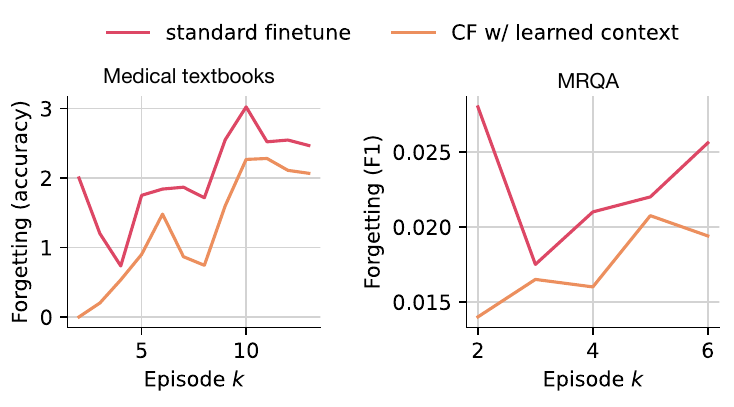}
    \vskip -0.05in
    \caption{Average forgetting $F_k$ over the entire course of continual learning. Conditional finetuning has consistent less forgetting than standard finetuning. (LLaMA-2 7B)}
    \label{fig:continual_forgetting}
\end{figure}

\cref{tab:continual_learning} shows that conditional finetuning has less forgetting and achieves better cumulative accuracy than standard finetuning, especially with learned context. \cref{fig:continual_forgetting} shows that conditional finetuning consistently has less forgetting of knowledge learned from previous corpora, over the entire course of continual learning.

In transfer learning where only a single context is used, the choice of context seems not affect performance much. In continual learning, the choice of context for each corpus can have a significant impact on performance. As we have shown in \cref{sec:conditional_lm}, the model learns a conditional topic prior $p(\text{topic}|a)$ in conditional finetuning. If a similar context is later used in training on a different corpus, the context will activate the previous learned topic prior which provides wrong information about the current corpus. 
As a result, the model may need to unlearn the conditional topic prior of the previous corpus before learning from the new corpus, hindering the learning of new knowledge.




Therefore, to achieve the maximum ``selectivity" in selective learning and reduce useless learning as much as possible, it is preferable to use contexts that provide information specific to each corpus when continually finetuning on multiple corpora.
To verify this relationship between context choice and performance, we measure the similarity between contexts under the three context types. We calculate the average pairwise KL-divergence between conditional data distributions given context $a_i$ ($a_i$ is used on the $i$-th corpus):
\begin{align}
    \frac{1}{n(n-1)}\sum_{i=1}^n \sum_{j=1, j\neq i}^n D_{KL}(p(x|a_i)||p(x|a_j))
\end{align}
The \textit{random} contexts have a average KL-divergence of 0.006, the \textit{domain hint} contexts 0.025, and the \textit{learned} contexts 0.033. This shows that the random UUIDs are semantically very similar, while the learned contexts are the most dissimilar. The learned contexts also provide most specific information about each corpus as they are optimized for each corpus. The similarity between contexts is inversely correlated with performance observed in \cref{tab:continual_learning}. 
The use of context specific to each corpus in conditional training leads to better selective learning and less forgetting.

\section{Discussion}

In this paper, we explored the effect of conditional finetuning, specifically on language modeling conditioned on a context. We found that by utilizing context to explain away corpus-level statistics, conditional training allows for selective learning from a corpus.  It allows learning knowledge useful for downstream tasks while minimizing the learning of useless corpus statistics, such as topic biases. As a result, conditional training reduces side effects of domain finetuning and achieves less forgetting. 

Selective learning gives finer-grained control over what the model learns in language modeling, and could be utilized for multiple purposes beyond discussed here. For example, it could help keeping the language model unbiased and better retain its general-purpose ability in continual finetuning and lifelong learning scenarios. For statistically biased corpora, 
conditional training may reduce the model's learning of the biases in sensitive attributes like race and gender, like in previous approaches \cite{productofexperts, productofexperts2}.

\paragraph*{Limitations.} We studied the effect of conditional finetuning on language models with a limited-size corpus due to computational limitations. It might require further investigation to check whether the effect of conditional finetuning scales to large-scale training (e.g., billions of tokens).

In terms of evaluation, we evaluate model's performance with QA tasks on the main concepts in the corpus, which evaluates model's memorization and basic understanding of the concepts in the corpus. We did not verify whether the model can apply the learned knowledge in more complex reasoning scenarios, which seems challenging for current language models \cite{kediteval, lmreverse}, and how conditional training affects such abilities.

Although conditional finetuning is observed to reduce forgetting, it is not proposed as a solution to catastrophic forgetting. Over-adaptation and unnecessary learning of corpus statistics is likely only one of the many factors that cause forgetting. We mainly aim to understand the effect of conditional learning in this paper, and leave the development of more effective methods to reduce forgetting to future work.

\section*{Impact Statement}

Our work mainly explores the effect of training language models with a context, and the results indicate that contexts can be used to exclude simple corpus statistics from learning by the model. This may be used to reduce social bias and improve the fairness of language models, because social bias is often simple bias caused by learning on a biased dataset. We have not yet foreseen any potential negative ethical consequences requiring particular discussion here.

\section*{Acknowledgements}
The work is supported by National Key R\&D Program of China (2021ZD0113402). We thank the anonymous reviewers for helpful comments and feedback.


\bibliography{bib_all}
\bibliographystyle{icml2024}

\newpage
\appendix
\onecolumn

\section{Data}

\subsection{Medical domain}
\label{app:medical}

\subsubsection{Corpus}
\label{app:corpus}

We use the medical textbooks provided with the MedQA dataset \cite{medqa} as a knowledge-rich corpus in the medical domain. To avoid varying the corpus size too much in continual learning setting, we use the 13 textbooks (1 subject each) that have size in the range of 1-10MB. 
\cref{tab:qa_stat} shows a statistics of the medical textbooks used. Number of tokens is measured with the tokenization scheme of LLaMA-2 \cite{llama2} model.

\subsubsection{QA task}
\label{app:qa}

We use GPT-4 to generate multiple-choice quiz questions on each medical subject. Given a medical subject, we use the procedure to generate questions:
\begin{enumerate}
    \item Split the textbook material of the subject into excerpts of 2048 tokens long, then randomly sample 50 excerpts (to help reduce the cost of GPT-4 usage). For each excerpt, 

    \item Instruct GPT-4 to generate 10 multiple-choice quiz questions examining the key concepts covered in the excerpt. The prompt is as follows:
    
    {\itshape Here is an excerpt from a {subject} textbook:\\

    \textlangle excerpt\textrangle\\
    \{input\}\\
    \textlangle/excerpt\textrangle\\
    
    Please write 10 multiple-choice quiz questions to examine whether a student remembers the key concepts from the above excerpt, after they studied the entire textbook.\\
    
    Requirements on content:\\
    - each question should have four choices, one choice must be definitely correct, the other three choices must be definitely wrong\\
    - the choices should be short and simple\\
    - each question should examine different key concepts in the material\\
    - provide enough context in the question so that it is answerable unambiguously, but do not refer to the particular excerpt, the figures, or the textbook\\
    - do not use negation (e.g., ``not", ``except") in the question, and do not use combination (e.g., ``all of the above", ``both A and B") in the choices\\
    
    Requirements on format:\\
    - please provide questions and answers in the following format: \\
    Question: \textlangle question\textrangle\\
    A) \textlangle choice 1\textrangle\\
    B) \textlangle choice 2\textrangle\\
    C) \textlangle choice 3\textrangle\\
    D) \textlangle choice 4\textrangle\\
    Answer: \textlangle the answer (a single letter)\textrangle\\
    - please directly give output without comments}

    where \{subject\} is replaced by the subject and \{input\} is replaced by the excerpt.

\end{enumerate}

\begin{table}[h]
    \centering
    \begin{tabular}{lcc}
    \toprule
    \textbf{Subject}  & \textbf{Corpus length (tokens)} & \textbf{\# Questions} \\ \midrule
    Anatomy & 661K & 500 \\
    Biochemistry & 404K & 500 \\
    Cell biology & 1296K & 500 \\
    Gynecology & 1768K & 500 \\
    Histology & 841K & 500 \\
    Immunology & 969K & 500 \\
    Neurology & 2272K & 500 \\
    Obstetrics & 2156K & 500 \\
    Pathology & 1122K & 500 \\
    Pediatrics & 842K & 500 \\
    Pharmacology & 1467K & 500 \\
    Physiology & 889K & 500 \\
    Psychiatry & 821K & 500 \\
    \bottomrule
    \end{tabular}
    \caption{Statistics of the medical textbook corpus for each subject.}
    \label{tab:qa_stat}
\end{table}

\cref{tab:qa_example} shows examples of the generated questions for some subjects.

\begin{table}[h]
    \centering
    \begin{tabular}{l}
    \toprule
    \textbf{Subject}  \\ \midrule
    \textit{Anatomy} \\
    Question: The inferior gluteal nerve innervates which of the following muscles?\\
    A) Tensor fasciae latae\\
    B) Gluteus medius\\
    C) Gluteus maximus\\
    D) Obturator internus\\
    Answer: C\\
    \midrule
    \textit{Biochemistry} \\
    Question: Which compound is an allosteric inhibitor of glutamate dehydrogenase (GDH)?\\
    A) Adenosine triphosphate\\
    B) Adenosine diphosphate\\
    C) Guanosine diphosphate\\
    D) Guanosine triphosphate\\
    Answer: D\\
    \midrule
    \textit{Cell biology} \\
    Question: In bacterial transcription, what helps the core enzyme break free from its interactions with promoter DNA?\\
    A) The binding of ribonucleotides\\
    B) Sigma factor reassociation\\
    C) Transcription bubble contraction\\
    D) Stress generated by scrunching\\
    Answer: D\\
    \midrule
    \textit{Gyneacology} \\
    Question: What is the recommended diagnostic step for premenarcheal patients with a pelvic mass?\\
    A) MRI scan\\
    B) Karyotype determination\\
    C) Pelvic ultrasound\\
    D) Hormone level testing\\
    Answer: B\\
    \midrule
    \textit{Histology} \\
    Question: In hepatocytes, where are lysosomes typically concentrated?\\
    A) Near the bile canaliculus\\
    B) Throughout the cytoplasm evenly\\
    C) Inside the nucleus\\
    D) At the cell periphery\\
    Answer: A\\
    \bottomrule
    \end{tabular}
    \caption{Examples of the generated questions, for the first 5 subjects in medical textbooks.}
    \label{tab:qa_example}
\end{table}

\subsection{General domain}

We use the MRQA benchmark \cite{mrqa} for evaluating knowledge learning on general domain. We extract all the documents from the 6 reading comprehension datasets as training corpora. The datasets are SQuAD \cite{squad}, NewsQA \cite{newsqa}, TriviaQA \cite{triviaqa}, SearchQA \cite{searchqa}, HotpotQA \cite{hotpotqa}, and Natural Questions \cite{naturalquestions}. To balance the size of the corpora and reduce computation cost, we sample 1,000 questions and the associating documents from each dataset. The questions are turned into a closed-book format QA, and the documents are used as the training corpus for finetuning the language models.

\section{Training details}

\subsection*{Language model finetuning}
\label{app:lm_finetuning}
We finetune the model with the AdamW optimizer \cite{adamw} with a learning rate of 3e-5. A linear learning rate decay is used with a warm-up of 10\% of the total number of steps. We use a gradient clipping at 1.0. Batch size is set to 16. The maximum sequence length is set to 2048. 

\subsection*{Prompt tuning}
\label{app:prompt_tuning}
To learn a soft prompt, we train the 10 embedding vectors on the training corpus for 3 epochs with a learning rate of 1e-1. The learning rate is chosen from a range search that minimizes loss. The training objective is conventional causal language modeling loss. 
The 10 embedding vectors have the same dimensionality as the language model's token embeddings.
Prompt tuning is performed with the PEFT \cite{peft} library. The learned soft prompts are fixed when used as a context in conditional finetuning.

\section{Evaluation details}
\label{app:qa_eval}

All QA tasks are evaluated using EleutherAI's Language Model Evaluation Harness framework \cite{lmevalharness}. 

The evaluation format for multiple choice-style QA tasks (anatomy, medical textbooks) follows the format of the MMLU \cite{mmlu} benchmark.

The evaluation format for completion-style QA tasks (SQuAD, MRQA) follows the evaluation protocol of the SQuAD \cite{squad} benchmark, which uses normalized F1 scores as metric. Answers and groudtruths are normalized and have articles and punctuation removed before word-level F1 score is calculated. The protocol is often used in a combined evaluation on multiple QA tasks \cite{decathlon, mrqa}.

All QA tasks are evaluated with 5-shot prompting.

For language modeling perplexity, when evaluating on C4, we randomly sampled 10,000 documents from the validation split of the English part of C4 corpus (C4/en) due to the large size of C4.





\begin{figure}[h]
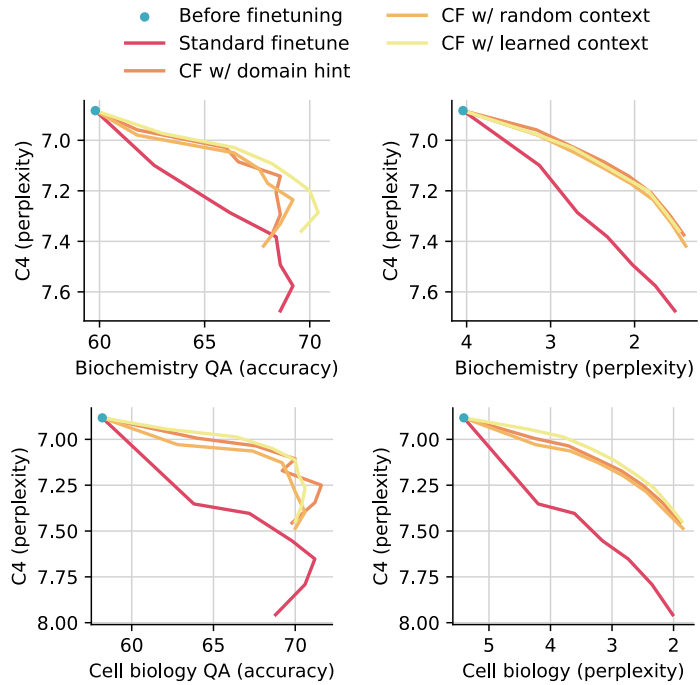

    \centering
    \includesvg[width=0.55\linewidth]{figures/transfer_biochemistry.svg}
    \includesvg[width=0.55\linewidth]{figures/transfer_cell_biology.svg}
    \caption{Performance-forgetting tradeoff curve of standard finetuning and conditional finetuning on Biochemistry and Cell biology, the second and third subjects in medical textbooks.
    Conditional finetuning has consistently less forgetting at similar levels of performance on downstream tasks, achieving significantly better tradeoff than standard finetuning.}
    \label{fig:qa_transfer_more}
\end{figure}

\end{document}